\documentclass{article}
\usepackage[utf8]{inputenc}
\usepackage{main}
\usepackage{microtype}
\usepackage{graphicx}
\usepackage{times}
\usepackage{amsmath}
\usepackage{enumitem}
\usepackage{booktabs}
\usepackage{xcolor}     
\usepackage{natbib}
\usepackage{url}
\usepackage{todonotes}
\usepackage{multirow}
\usepackage{makecell}
\usepackage{color-edits}
\usepackage{siunitx}
\usepackage{subcaption}
\usepackage{sidecap}
\usepackage{float}

\definecolor{mydarkblue}{rgb}{0,0.08,0.45}
\usepackage[colorlinks=true,allcolors=perfblue]{hyperref} 

\usepackage[customcolors]{hf-tikz}
\definecolor{expert}{HTML}{008000}
\definecolor{error}{HTML}{f96565}
\definecolor{learner}{HTML}{F79646}
\definecolor{perfblue}{RGB}{64, 114, 175}

\usepackage{pifont}
\newcommand{\cmark}{\ding{51}}%
\newcommand{\xmark}{\ding{55}}%
\usepackage{float}

\usepackage{amsfonts, amsmath, amssymb, bm}       %
\usepackage{amsthm}
\usepackage{thmtools}

\usepackage{fancyhdr}

\definecolor{gred}{RGB}{250, 210, 207}
\definecolor{coolblue1}{rgb}{0.91, 0.94, 0.98}
\definecolor{coolblue2}{rgb}{0.76, 0.85, 0.94}
\definecolor{coolblue3}{rgb}{0.54, 0.72, 0.87}
\definecolor{coolblue4}{rgb}{1, 1, 1}

\usepackage{xspace}
\usepackage{cleveref}

\newenvironment{itemize*}%
 {\leftmargini=10pt\begin{itemize}%
  \setlength{\itemsep}{0pt}%
  \setlength{\parskip}{0pt}%
  }%
 {\end{itemize}}
\newenvironment{enumerate*}%
 {\begin{enumerate}%
  \setlength{\itemsep}{0pt}%
  \setlength{\parskip}{0pt}}%
 {\end{enumerate}}

\begin{document}

\title{\textbf{\textit{Gained In Translation}: \\Privileged Pairwise Judges Enhance Multilingual Reasoning}}

\author{
\textbf{Lintang Sutawika}$^{1}$ \quad
\textbf{Gokul Swamy}$^{2}$ \quad
\textbf{Zhiwei Steven Wu}$^{3}$ \quad
\textbf{Graham Neubig}$^{1}$ \\
\textsuperscript{1}Carnegie Mellon University, Language Technologies Institute \\
\textsuperscript{2}Carnegie Mellon University, Robotics Institute \\
\textsuperscript{3}Carnegie Mellon University, Software and Societal Systems Department \\
\texttt{\{lsutawik, gswamy, zstevenwu, gneubig\}@cs.cmu.edu} \\\\
\href{https://github.com/lintangsutawika/sp3f}{\includegraphics[height=0.4cm]{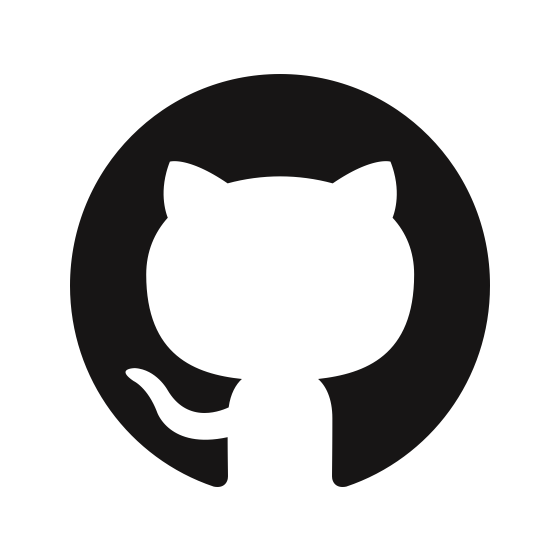} {\textbf{lintangsutawika/sp3f}} ~ ~ ~ \href{https://huggingface.co/collections/neulab/sp3f}{\includegraphics[height=0.4cm]{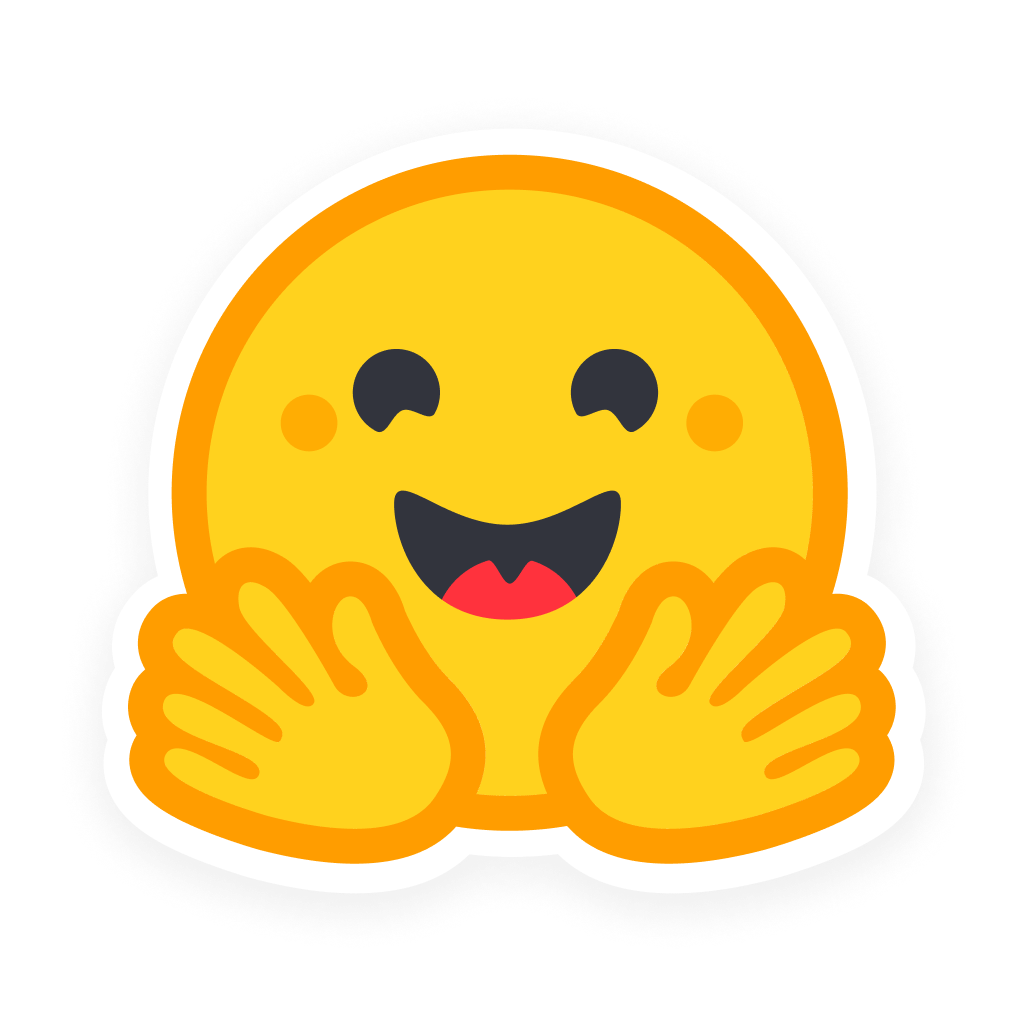} \textbf{neulab/sp3f}}  
}
}

\maketitle
\thispagestyle{fancy}
\fancyhead{}
\lhead{\includegraphics[height=0.5cm]{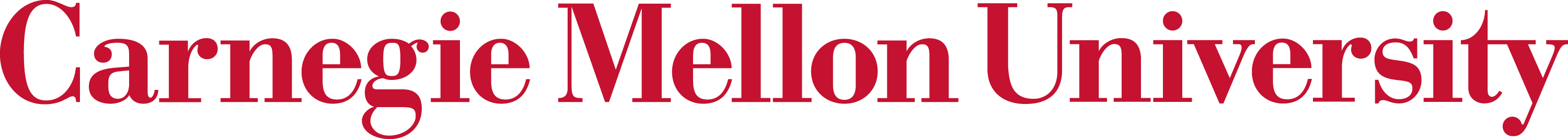}}
\rhead{%
  \raisebox{-0.1cm}{\includegraphics[height=0.8cm]{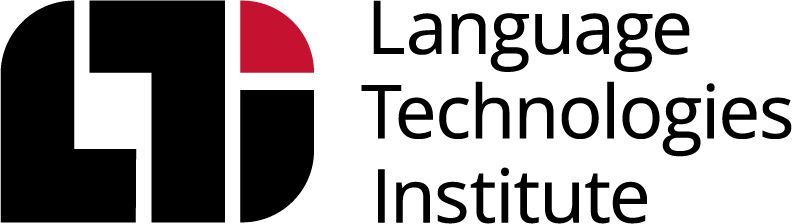}}%
}
\renewcommand{\headrulewidth}{0pt}
\setlength{\headheight}{12pt}
\addtolength{\topmargin}{00pt}
\setlength{\headsep}{3mm}

\vspace{-1.0em}

\begin{abstract}
When asked a question in a language less seen in its training data, current reasoning large language models (RLMs) often exhibit dramatically lower performance than when asked the same question in English.
In response, we introduce \texttt{SP3F} (Self-Play with Privileged Pairwise Feedback), a two-stage framework for enhancing multilingual reasoning without \textit{any} data in the target language(s). First, we supervise fine-tune (SFT) on translated versions of English question-answer pairs to raise base model correctness. Second, we perform RL with feedback from a pairwise judge in a self-play fashion \cite{swamy2024minimaximalistapproachreinforcementlearning}, with the judge receiving the English reference response as \textit{privileged information}. Thus, even when none of the model's responses are completely correct, the privileged pairwise judge can still tell which response is better. End-to-end, \texttt{SP3F} greatly improves base model performance, even outperforming fully post-trained models on multiple math and non-math tasks with less than $1/8$ of the training data across the single-language, multilingual, and generalization to unseen language settings.
\end{abstract}

\begin{figure}[h]
    \centering
    \includegraphics[width=0.5\linewidth]{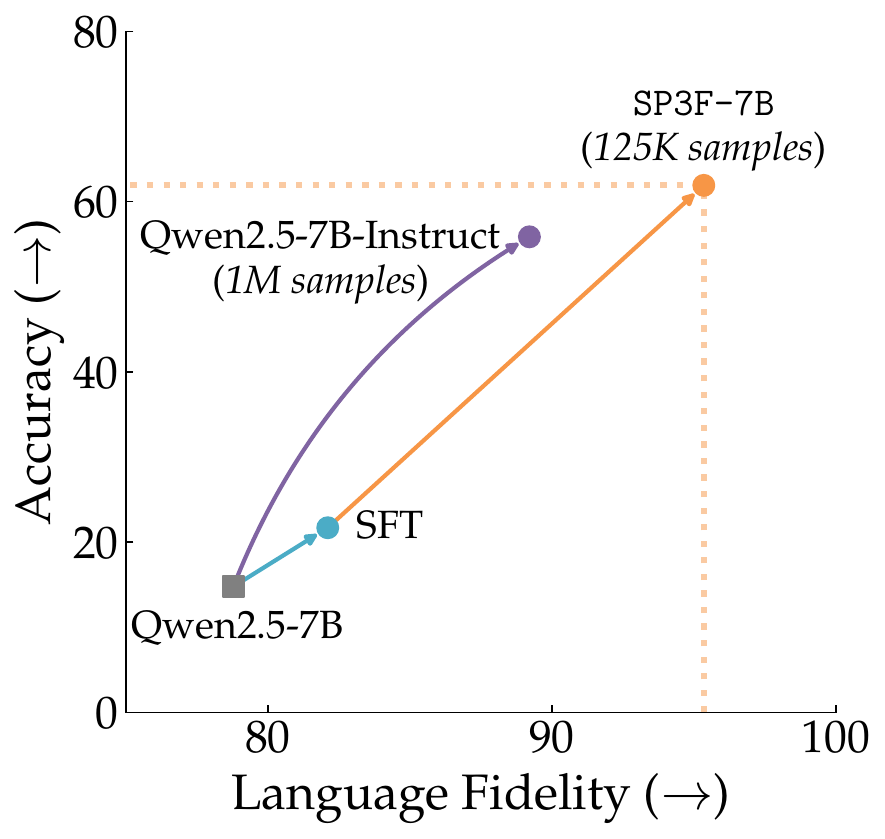}
    \caption{
        We propose \texttt{SP3F}: \textit{Self-Play with Privileged Pairwise Feedback}: a method for training multilingual reasoning models without \textit{any} data in the target language(s). \texttt{SP3F-7B} out-performs Qwen2.5-7B-Instruct across 4 tasks with roughly $1/8$ of the training data ($125,000$ for \texttt{SP3F-7B} vs. $1,000,000$ for Qwen2.5-7B-Instruct), both in terms of accuracy and language fidelity (did the model answer in the target language?).
        }
    \label{fig:ffig}
\end{figure}

\twocolumn
\begin{figure*}[h]
    \centering
    \includegraphics[width=0.9\linewidth]{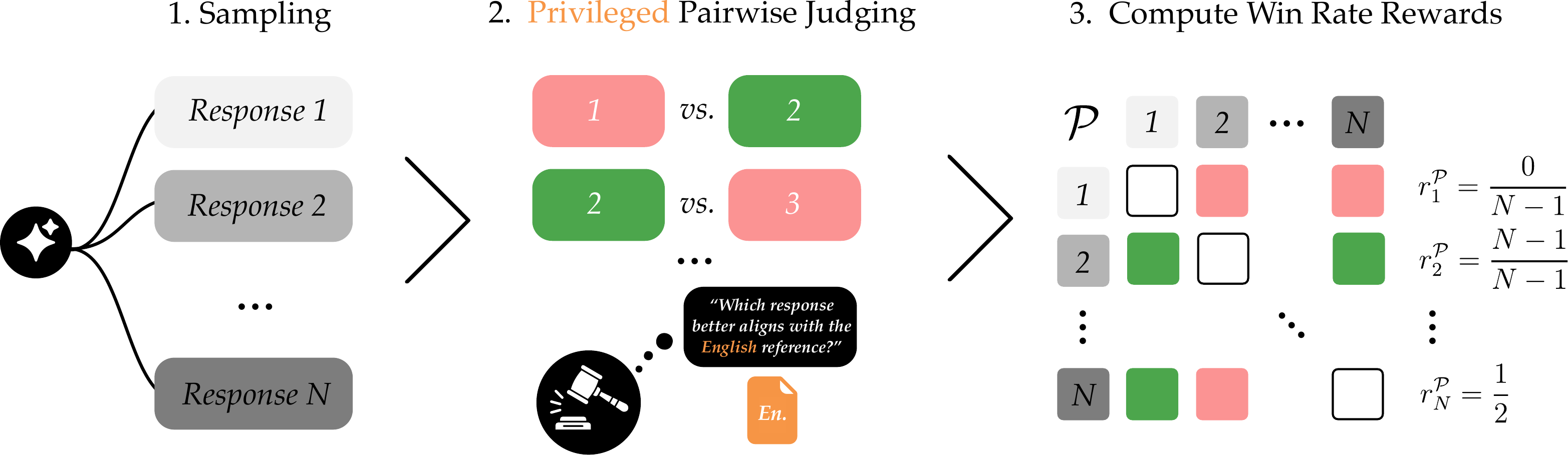}
    \caption{
    The second stage of the \texttt{SP3F} pipeline is to perform RL (GRPO, \citet{shao2024deepseekmathpushinglimitsmathematical}) with feedback from verifiable rewards \cite{lambert2025tulu3pushingfrontiers} and a pairwise judge. To aid in its judgments, the judge LLM is given access to
 \textit{privileged information} in the form of an English reference response. Concretely, we sample $N$ responses from the model (left), ask the privileged judge to pick a winner from each pair (center), and then use the average win-rate of each response against the other $N-1$ samples as the reward for RL (right, \citet{swamy2024minimaximalistapproachreinforcementlearning}).
    }
    \label{fig:full_pipeline}
\end{figure*}

\section{Introduction}

Current reasoning large language models (RLMs) are trained on data (e.g., chains of thought, CoTs) that is primarily in English \cite{ghosh2025surveymultilingualreasoninglanguage}. This means that when an RLM is asked the same question in a non-English language, it often exhibits dramatically lower performance than if it were asked the question in English \citep{yong2025crosslingualreasoningtesttimescaling, muennighoff2023crosslingualgeneralizationmultitaskfinetuning, shi2022languagemodelsmultilingualchainofthought,tam2025languagemattersmultilingualinput}.

Improving reasoning performance in \textit{lower resource} languages (e.g., Indonesian, Swahili, Bengali) is challenging as we lack large amounts of data in the target language for supervised fine-tuning (SFT), and the base model's probability of generating the correct answer might be so low that getting positive signal for reinforcement learning (RL) to succeed is computationally challenging. Furthermore, for reasoning tasks, outcome-level \textit{verifiable rewards} \citep{lambert2025tulu3pushingfrontiers} that consider just the final answer provide only indirect supervision on the CoT, making exploration challenging due to the sparsity of feedback \citep{kakade2003sample}. Put together, we face a \textit{cold start} problem we can't easily offline fine-tune our way out of.

In response, we propose \texttt{SP3F} (Self-Play with Privileged Pairwise Feedback): a two-stage framework for increasing reasoning performance in non-English target language(s) that doesn't require \textit{any} data in the target language(s). First, we apply SFT on \textit{translated} versions of English reference responses to raise our RLM's probability of generating correct answers. Second, we perform RL with a combination of verifiable rewards (e.g., answer correctness, language fidelity) and preference feedback from an LLM judge \citep{zheng2023judgingllmasajudgemtbenchchatbot}. The LLM judge directly supervises the CoT of the RLM, giving it a direction of improvement even when it can't produce a correct final answer, which helps mitigate the cold start issue of lower-resource language reasoning.

While conceptually promising, the noisiness of the feedback provided by LLM judges makes incorporating them into RL training challenging. First, if the LLM judge itself is unfamiliar with a lower-resource language, it may be unable to provide accurate feedback. In response, we provide the English reference response as \textit{privileged information} \cite{vapnik2009new} to the judge, asking it to merely pick which of the two RLM responses more closely aligns with the English reference response. This is an easier \emph{translation} task than judgment in the abstract. We find that the use of privileged information improves judgment quality.

Second, due to their pretraining on vast swathes of internet text, LLMs often exhibit \textit{intransitive} (i.e., cyclic) preferences where they might rank $A \succ B$, $B \succ C$, and $C \succ A$ \citep{xu2025investigatingnontransitivityllmasajudge}. Such intransitivity means no scalar reward function can faithfully represent the judge's preferences, making standard reward modeling fundamentally misspecified. Rather than fitting an inconsistent reward model, we adopt a \emph{self-play} style approach that optimizes pairwise preferences directly: after sampling a batch of candidate responses, we use the judge to compare all pairs and assign each response a score equal to its empirical win rate. This aggregation converts pairwise judgments into a learning objective that reliably improves the model despite intransitivity \citep{swamy2024minimaximalistapproachreinforcementlearning}.  Put together, we propose to use \textit{privileged pairwise judges} to provide denser feedback to the RLM during RL.

Our key insight is that \textbf{\textit{we can use English reference responses during both SFT and RL by framing both learning problems in terms of translation.}} We use reference responses as data for translation during SFT and as privileged information for the pairwise judge during downstream RL. 
Our contribution is three-fold:

\begin{enumerate}

\item \textbf{We introduce \texttt{SP3F}: a multi-step framework for increasing reasoning performance in a target language without data in said language.} We find that RLMs trained via \texttt{SP3F} out-perform fully post-trained models on both in-domain math and out-of-domain non-math tasks in a target language.

\item \textbf{We apply \texttt{SP3F} on data from 18 languages, producing a model that out-performs fully post-trained models using $\frac{1}{8}$ as much training data.} We outperform Qwen2.5-7B-Instruct across math and on-math reasoning tasks. We find particularly large improvements on lower-resource languages and see better generalization to unseen languages. %

\item \textbf{We perform an in-depth exploration of the benefits provided by privileged information.} We find that privileged information is particularly helpful with reducing the intransitivity of the judge model, as well as in improving detection of correct reasoning chains, even if the final answer is incorrect.

\end{enumerate}

\begin{table*}[t]
\centering
\setlength{\tabcolsep}{5pt}
\scalebox{0.85}{
\begin{tabular}{l|cc|cccc|cccccc}
\toprule
\multirow{1}{*}{Model}
 & \multicolumn{2}{c}{Overall} & \multicolumn{2}{c}{MGSM} & \multicolumn{2}{c}{MT Math100} & \multicolumn{2}{c}{Belebele} & \multicolumn{2}{c}{\makecell{Global MMLU \\ Lite}}  \\
 & Acc & Lang & Acc & Lang & Acc & Lang & Acc & Lang & Acc & Lang  \\
\midrule
Qwen2.5-7B & 14.79 & 78.78 & 22.15 & 90.67 & 21.16 & 58.22 & 7.52 & 80.39 & 8.34 & 85.85 \\
\quad + SFT & 21.70 & 82.11 & 33.66 & 91.37 & 26.72 & 58.26 & 12.94 & 89.18 & 13.48 & 89.62 \\
\quad\quad + RLVR & \underline{57.79} & \textbf{96.09} & 65.34 & \textbf{99.75} & 44.50 & \textbf{86.10} & \textbf{68.18} & \underline{98.73} & \underline{53.15} & \textbf{99.78} \\
\texttt{SP3F-7B} & \textbf{61.91} & \underline{95.35} & \textbf{72.50} & \underline{99.38} & \underline{56.84} & \underline{82.93} & \underline{67.54} & \textbf{99.65} & 50.76 & \underline{99.45} \\
\midrule
Qwen2.5-7B-Instruct & 55.87 & 89.21 & \underline{66.36} & 98.38 & 52.12 & 65.66 & 56.79 & 96.59 & 48.20 & 96.21 \\
\quad + Translate Test & 57.01 & 85.98 & 66.15 & 95.81 & \textbf{60.08} & 59.34 & 48.09 & 92.27 & \textbf{53.73} & 96.49 \\
\bottomrule
\end{tabular}
 }
\caption{Across in-domain math tasks (MGSM and MT Math100) and out-of-domain tasks non-math tasks (Belebele and Global MMLU Lite) over a subset of 18 languages (\autoref{tab:language_list}) that were used to train \texttt{SP3F-7B}, we see \texttt{SP3F-7B} consistently outperforms the Qwen2.5-7B-Instruct. We measure performance in percentage by Accuracy (Acc) and Language Fidelity (Lang). Highest score presented in \textbf{bold} and second highest \underline{underlined}. Notably,  \texttt{SP3F-7B} required only $\frac{1}{8}$ as much data to post-train Qwen2.5-7B-Instruct. Full results in Appendix \ref{sec:full_table_results}.
}
\label{tab:all_tasks}
\end{table*}

\section{\texttt{SP3F}: Self-Play with Privileged Pairwise Feedback for Multilingual Reasoning}
\label{sec:method}

In this section, we begin by describing \texttt{SP3F} in detail. \texttt{SP3F} is a two-step framework for improving reasoning performance in a target language without data in said language.
\texttt{SP3F} only requires English reference responses, which can be relatively easily generated by a teacher model (e.g., o1 \citep{jaech2024openai}, R1 \cite{deepseekai2025deepseekr1incentivizingreasoningcapability}).

Below, we use $x \in \mathcal{X}$ to refer to prompts/questions and $y \in \mathcal{Y}$ to refer to responses, with $y^{\star}$ referring to an (English) reference response. We assume access to a dataset $\mathcal{D}$ of $(x, y^{\star})$ pairs. Each response $y$ consists of a chain-of-thought $z \in \mathcal{Z}$ and response $a \in \mathcal{A}$ (i.e., $y = (z, a) \in \mathcal{Y} = \mathcal{Z} \times \mathcal{A}$). We search over policies $\pi \in \Pi \subseteq \{\mathcal{X} \to \Delta(\mathcal{Y})\}$. We use $\circ$ to denote the concatenation of two strings and $\mathsf{tx}(\cdot)$ to denote translation into the appropriate target language. There are two stages of the \texttt{SP3F} pipeline: an SFT stage, followed by an RL stage.

\noindent \textbf{Stage 1: SFT on Translated English Responses.} Ideally, we would solve the cold-start problem of reasoning in a lower-resource language by training on data in the target language. However, by definition, there is a relatively limited about of data available in a target language. Furthermore, it is often difficult to learn a strong policy via SFT given limited amounts of training data \citep{swamy2025roadsleadlikelihoodvalue}.

We propose a simple solution to this problem: SFT \textit{translations} of relatively plentiful English reference responses $(x, y^{\star})$.  Explicitly, we maximize likelihood via a standard next-token prediction loss:
\begin{equation}
    \pi_{\mathsf{sft}} = \arg\max_{\pi \in \Pi} \sum_{i=1}^{|\mathcal{D}|} \log \left(\pi(\mathsf{tx}(y^{\star}_i)\vert \mathsf{tx}(x_i))\right).
\end{equation}
Performing this process raises our RLM's probability of generating the correct answer in the target language, aiding in downstream mode selection via online RL \cite{yue2025doesreinforcementlearningreally} in the next stage.

\noindent \textbf{Stage 2: RL with Verifiable Rewards $+$ Privileged Pairwise Judge Feedback.}
Next, we perform RL, with rewards given via a composition of four terms: three verifiable binary indicators, and one batch-level judge feedback term. Explicitly, given $N$ responses $y_{1:N} \sim \pi(x)$, we compute:
\begin{align}
    r(x, y_i, y^{\star}) &= r^{\mathsf{acc}}(x, y_i) + r^{\mathsf{fmt}}(y_i) + r^{\mathsf{lang}}(y_i) \nonumber \\ &+ r^{\mathcal{P}}(x, y_i, y_{1:N}, y^{\star}).
\end{align}

\noindent \textbf{\textit{Verifiable Rewards.}} The first three terms, $r^{\mathsf{acc}}(x, y_i) \in \{0, 1\}$ (accuracy), $r^{\mathsf{fmt}}(y_i) \in \{0, 1\}$ (formatting), and $r^{\mathsf{lang}}(y_i) \in \{0, 1\}$ (language fidelity), are each verifiable rewards. In particular, $r^{\mathsf{acc}}(x, y_i)$ measures if the answer $a_i$ is correct, $r^{\mathsf{fmt}}(y_i)$ measures if answer $a_i$ was provided inside a \texttt{\textbackslash boxed\{\}} template, and $r^{\mathsf{lang}}(y_i)$ measures whether the response was indeed in the target language. We use an automated language classifier to check what fraction of the response is in the target language. If the fraction $\geq 70\%$, we output a score of 1. We found that providing a binary indicator of the target language content, rather than a scalar in $[0, 1]$, helped avoid ``reward-hacking'' \citep{hadfield2017inverse}, where the model would learn to output a short response to achieve $100\%$ language fidelity. We chose $70\%$ as our threshold to account for the fact that math symbols are not counted as part of any particular language.

\noindent \textit{\textbf{Judge Feedback Reward.}} Even after SFT, our RLM may still have a relatively low probability of generating a correct reasoning chain. As the verifiable rewards only focus on the correctness of the answer and the language fidelity of the CoT, they do not provide direct supervision on the the correctness of the CoT. In response, we propose using an LLM judge for supervision on the CoT $z$. This can provide a clear direction of improvement even when the RLM can't answer the question completely correctly. We use gpt-4o-mini as a judge . %

There are two key challenges with learning from LLM judge feedback for multilingual reasoning. The first is that the judge may struggle to evaluate responses in a lower resource language that it doesn't understand well itself. In response, we propose to give the judge (but not the RLM) access to \textit{privileged information} \citep{vapnik2009new} in the form of the English reference answer $y^{\star}$. Thus, rather than having to judge the model's response $y$ in the abstract, the judge merely needs to assess how closely the solution $y$ aligns with the English reference $y^\star$. This is an easier \emph{translation}-style task. From another angle, we're \textit{recycling} the data used during offline SFT during online RL, effectively squeezing more juice out of the same samples. This bears similarity to the work of \citet{jain2025smooth}.%
\footnote{In greater detail, when given privileged information, the judge is able to provide more accurate feedback without requiring parameter updates/training data. Thus, compared to non-privileged judges, we have reduced the \textit{sample complexity} of learning a \textit{verifier}. This directly translates to a reduction in the end-to-end sample complexity of learning a policy/\textit{generator} via the arguments presented in \citet{swamy2025roadsleadlikelihoodvalue}.}

The second challenge is that due to their pretraining on a wide variety of text scraped from the internet, LLM judges often exhibit \textit{intransitive} preferences \citep{xu2025investigatingnontransitivityllmasajudge}, where they might rank responses $y_A, y_B, y_C \sim \pi(x)$ as $y_A \succ y_B$, $y_B \succ y_C$, and $y_C \succ y_A$. We find significant intransitivity in our LLM judge, as we explore below. Such inconsistent feedback can be a challenge to learn from. In response, given $N$ samples in a batch, we use a \textit{pairwise judge} to pick a winner from each of the $\binom{N}{2}$ pairs and use the win rate of each sample as the reward. Such a \textit{self-play} approach is provably robust to intransitive preferences \citep{swamy2024minimaximalistapproachreinforcementlearning}. Put together, we optimize:
\begin{equation}
    r^{\mathcal{P}}(x, y_i, y_{1:N}, y^{\star}) = \sum_{j \neq i}^N \frac{\widetilde{\mathcal{P}}(y_i \succ y_j|x, y^{\star})}{N-1}, \label{eq:r-judge}
\end{equation}
where $\widetilde{\mathcal{P}}(y_i \succ y_j \vert x, y^{\star}) \in [0, 1]$, with a value of $1$ denoting that $y_i$ was preferred to $y_j$ by the privileged pairwise judge. To account for the positional bias of pairwise judges \citep{zheng2023judgingllmasajudgemtbenchchatbot, qin2024large}, we perform the standard averaging of judge preferences across both input orderings:
\begin{align}
    & \widetilde{\mathcal{P}}(y_i \succ y_j \vert x, y^{\star}) = \nonumber \\ &\frac{\mathcal{P}(y_i \succ y_j \vert x, y^{\star}) + (1 - \mathcal{P}(y_j \succ y_i \vert x, y^{\star}))}{2}. 
\end{align}

\noindent \textit{\textbf{RL Algorithm.}} We use the DR.GRPO \citep{liu2025understandingr1zeroliketrainingcritical} variant of GRPO \citep{shao2024deepseekmathpushinglimitsmathematical}. While industry standard practice for RLM training, these RL algorithms are particularly prone to mode selection \citep{shao2025spuriousrewardsrethinkingtraining, oertell2025heuristics}, underscoring the need for a preliminary SFT step.

\noindent \textit{\textbf{Multilingual Training.}} Finally, to take advantage of the repeatedly observed benefits of multilingual training \citep{yong2025crosslingualreasoningtesttimescaling,shi2022languagemodelsmultilingualchainofthought,muennighoff2023crosslingualgeneralizationmultitaskfinetuning}, we apply the above pipeline with data from 18 different languages (see \autoref{tab:language_list} for full list). We refer to the model that results from this process as \texttt{SPF3-7B}.

\begin{figure*}[ht]
    \centering
    \includegraphics[width=0.85\linewidth]{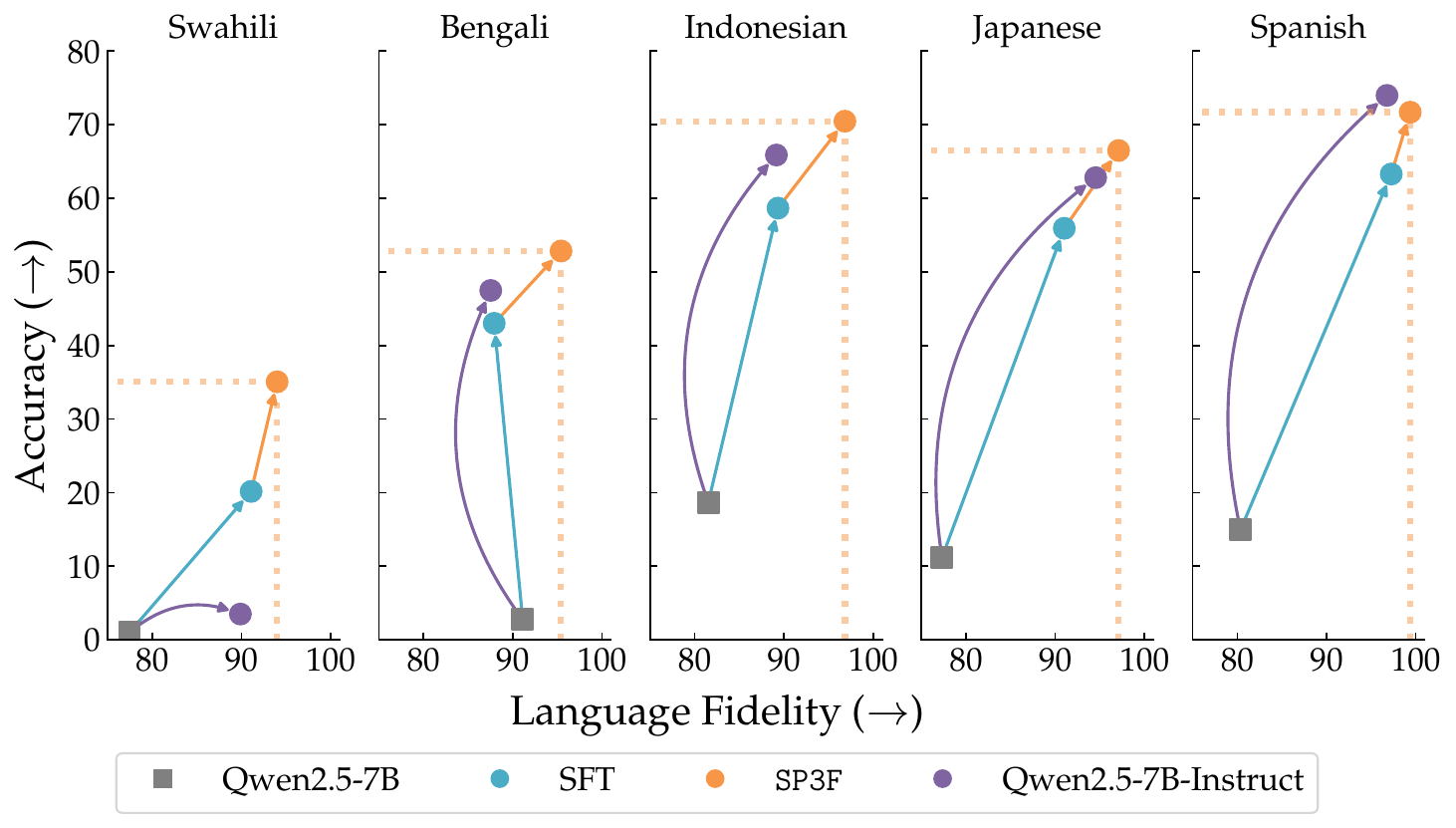}
    \caption{Across 5 target languages (ordered in terms of increasing resource tier), single-language \texttt{SP3F} training produces models that out-performs Qwen2.5-7B-Instruct. We find particularly large deltas on lower resource languages like Indonesian, Bengali, and Swahili. Results are averaged across all four tasks considered.}
    \label{fig:language_specific}
\end{figure*}

\section{Experiment Setup}
\label{sec:setup}

We now outline our experimental setup.

\noindent \textbf{Dataset Construction.} Our training data is generated from DeepScaleR~\citep{deepscaler2025}, which contains math reasoning problems from the AIME, AMC, and other competitions. We translate both the query and response to 18 languages (full list in \autoref{tab:language_list}) using GPT-5-Nano. All translated versions are then merged into a single dataset of equal proportions. Each sample includes the original English response, which is provided as privileged information.

\noindent \textbf{Training.} We train \texttt{SP3F-7B} on top of the Qwen2.5-7B~\citep{qwen2025qwen25technicalreport} base model. 
Our experiments were implemented in Verl~\cite{sheng2024hybridflow}, with a modification to allow for pairwise judges during reward calculation. For the SFT stage, we perform 1000 gradient steps with a batch size of 16 and learning rate of $\num{1e-5}$. For the secondary RL stage, we use a combination of verifiable rewards and judge feedback for supervision. We train for 500 gradient steps using a batch size of 32 prompts, $N=8$ responses per prompt, and a learning rate of $\num{5e-7}$.

\noindent \textbf{Metrics.} We report two metrics: \textit{Accuracy} (the correctness of the model's final \texttt{\textbackslash boxed\{\}} answer) and \textit{Language Fidelity} (whether the model's response is at least 70\% in the target language). To measure language fidelity, we use lingua.\footnote{https://github.com/pemistahl/lingua-py} These metrics are precisely the $r^{\mathsf{acc}}$ and $r^{\mathsf{lang}}$ discussed in Sec. \ref{sec:method}.

\noindent \textbf{Evaluation.} We evaluate all models on 2 math tasks and 2 non-math tasks. In the context of our math reasoning-based training data, the math tasks are in-domain, while the non-math tasks are out-of-domain tasks meant to provide an estimate of how well the model generalizes. For math tasks, we use MGSM~\citep{shi2022languagemodelsmultilingualchainofthought} to test basic word math problems and
MT-Math100~\citep{son2025linguistic} that is a translated subset of MATH500~\citep{lightman2023letsverifystepstep}.%
\footnote{We additionally contribute a new, translated and manually verified (by an author) Indonesian version of MGSM.}
For non-math tasks, we use Global MMLU Lite~\citep{singh2024globalmmluunderstandingaddressing} to evaluate world knowledge and Belebele~\citep{bandarkar-etal-2024-belebele} for reading comprehension. Each reported score is a per-question/prompt average over 8 model responses.

\noindent \textbf{Baselines.}
To compare against a strong post-training baseline, we choose Qwen2.5-7B-Instruct~\cite{qwen2025qwen25technicalreport}, a model post-trained by the Qwen team on 1M post-training examples.
In addition, we compare against Translate-Test~\cite{ponti2021modellinglatenttranslationscrosslingual, artetxe2023revisitingmachinetranslationcrosslingual}, where the query is translated into English before being solved by the model.
Specifically, we use Self-Translate Test~\cite{etxaniz2023multilinguallanguagemodelsthink} where the translation are done using the model itself. This technique is a training-free procedure to boost model reasoning performance.

\noindent \textbf{Using Privileged Information.} We use GPT-4o-mini as our LLM judge and provide it with the query in the target response language. Our system and user prompts instruct the model to deliberate over the responses A and B and decide which among them have closest sense to the included English reference response (full prompt available in \autoref{tab:privileged_info_prompts}). The judge is then directed to provide its final answer or either \texttt{\textbackslash boxed\{A\}} or \texttt{\textbackslash boxed\{B\}}.

\section{\texttt{SP3F} Unlocks Data-Efficient Multilingual Reasoning}

We begin by discussing the single-language gains of the \texttt{SP3F} pipeline, before discussing our multilingual results and more carefully exploring the benefits of privileged information for LLM judges.

\subsection{\texttt{SP3F} Improves Lower-Resource Language Reasoning}
As seen in \autoref{fig:language_specific}, applying the \texttt{SP3F} pipeline consistently boosts model performance above Qwen2.5-7B-Instruct level, averaged across both in-domain math tasks and out-of-domain non-math tasks. We see particularly large gains on the left side of the figure in lower-resource languages (e.g., Swahili, where the \texttt{SP3F}-trained model achieves more than three times the accuracy of Qwen2.5-7B-Instruct). Furthermore, we see that both stages of the \texttt{SP3F} pipeline are critical for strong final model performance. We emphasize that single-language \texttt{SP3F} uses significantly less data than the entire Qwen2.5-7B-Instruct post-training pipeline and no data in the target language.

\subsection{\texttt{SP3F} Improves Multilingual Reasoning}
We now explore the performance of \texttt{SP3F-7B}, which is trained by applying the \texttt{SP3F} pipeline on multilingual training data from 18 languages.

\noindent \textbf{Aggregate Results.} As seen in \autoref{tab:all_tasks}, \texttt{SP3F-7B} consistently out-performs Qwen2.5-7B-Instruct on both in-domain math tasks and out-of-domain not math tasks. This is impressive given \texttt{SP3F-7B} only required $\frac{1}{8}$ as much post-training data. Furthermore, even when we apply the inference-time translate test technique to Qwen2.5-7B-Instruct (but not our model), \texttt{SP3F-7B} still usually beats the improved model.

By comparing the +RLVR and \texttt{SP3F-7B} rows of \autoref{tab:all_tasks}, we can more precisely identify the benefits of judge feedback. We see that judge feedback improves accuracy at the cost of slightly worse language fidelity. Zooming in further, we see particularly strong gains on in-domain math tasks, with a slight decrease in performance relative to RLVR-trained models on out-of-domain non-math tasks. This suggests that better regularization techniques (e.g., those proposed by \citet{NEURIPS2024_16c628ab}) may be poised to improve the out-of-domain generalization of judge-trained models.

\begin{figure}[ht!]
    \centering
    \includegraphics[width=1.0\linewidth]{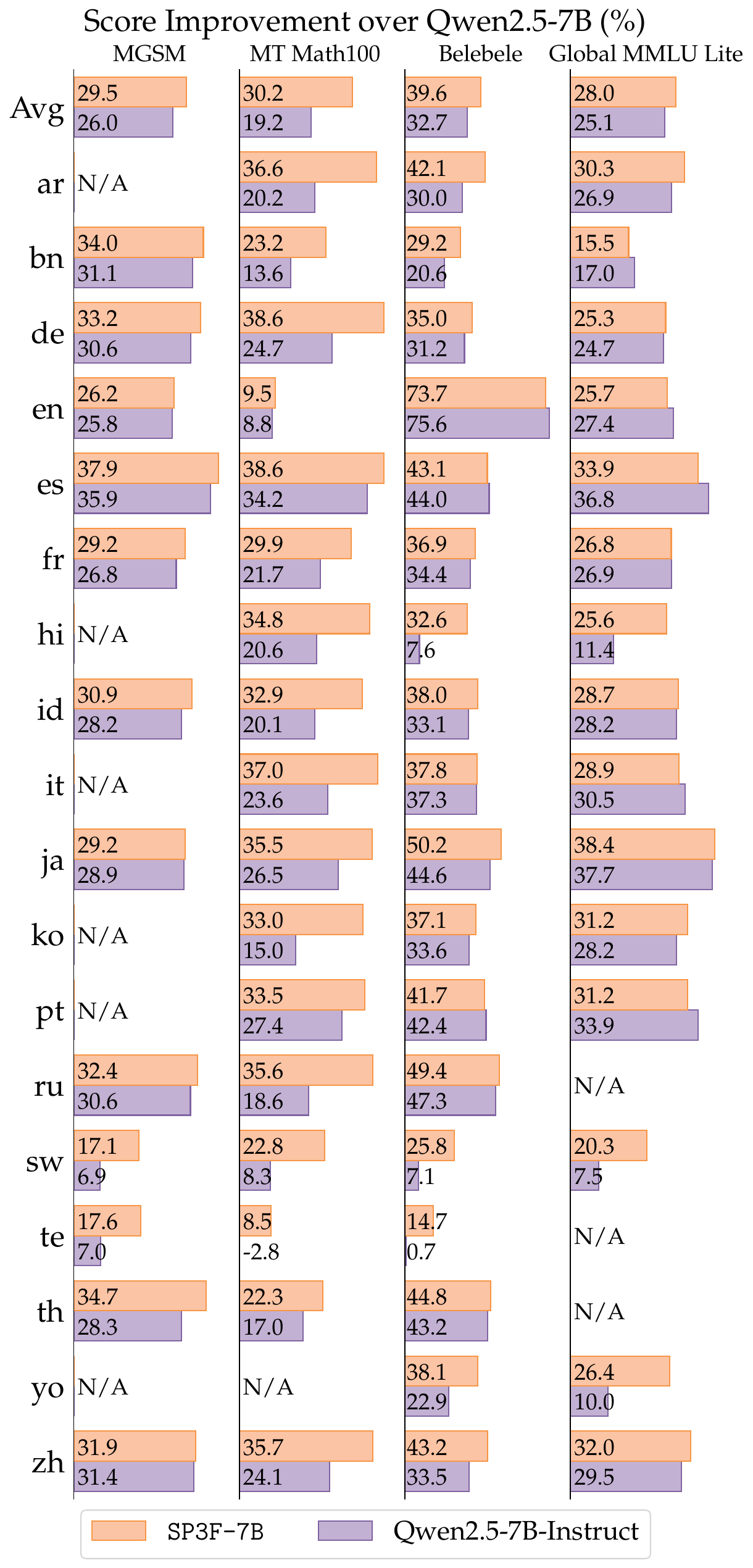}
    \caption{\texttt{SP3F-7B} generally outperforms Qwen2.5-7B-Instruct across most languages tested. We see particularly large gains for in-domain math tasks (left two columns) and on out-of-domain tasks in lower-resource languages (e.g., Swahili). Each bar represents the gain in terms of absolute performance points compared to Qwen2.5-7B for a specific language. N/A denotes that the task is not available for that language.}
    \label{fig:gain}
\end{figure}

\noindent \textbf{Per-Language Results.} In \autoref{fig:gain}, we display the gains over the base model broken down by target language. Echoing the results in \autoref{tab:all_tasks}, we see positive deltas over Qwen2.5-7B-Instruct on all languages on in-domain tasks. We also see particularly large positive deltas on out-of-domain tasks in lower-resource languages like Swahili, Hindi, Yoruba, and Telegu. 

\clearpage
\noindent \textbf{Unseen Language Generalization Results.} When we evaluate \texttt{SP3F-7B} on eight languages outside of its training set (but not necessarily that of Qwen2.5-7B-Instruct), we see better performance across tasks compared to Qwen2.5-7B-Instruct. As seen in \autoref{tab:unseen_lang}, with the exception of Gujarati (gu), \texttt{SP3F-7B} out-performs Qwen2.5-7B-Instruct by around $18\%$ on Belebele and $3.4\%$ on MT Math100 in terms of accuracy. This indicates that \texttt{SP3F-7B} is a more generally capable multilingual reasoning model than Qwen2.5-7B-Instruct and that the \texttt{SP3F} pipeline doesn't preclude generalization.

\begin{table}[h]
\centering
\resizebox{1.0\linewidth}{!}{
\begin{tabular}{c|cc|cc||cc|cc}
\toprule
 & \multicolumn{4}{c}{Belebele} & \multicolumn{4}{c}{MT Math100} \\
 & \multicolumn{2}{c}{Q2.5-7B-I} & \multicolumn{2}{c}{\texttt{SP3F-7B}} & \multicolumn{2}{c}{Q2.5-7B-I} & \multicolumn{2}{c}{\texttt{SP3F-7B}} \\
 & Acc & Lang & Acc & Lang & Acc & Lang & Acc & Lang \\
\midrule
Avg & 39.9 & 93.4 & \textbf{58.3} & \textbf{99.9} & 48.3 & 62.7 & \textbf{51.7} & \textbf{83.6} \\
\midrule
af & 19.2 & 98.8 & \textbf{72.5} & \textbf{100.0} & 55.9 & 62.9 & \textbf{59.1} & \textbf{83.7} \\
gu & \textbf{30.6} & 99.5 & 19.2 & \textbf{100.0} & 35.5 & 65.0 & \textbf{38.0} & \textbf{68.9} \\
he & 61.3 & 72.5 & \textbf{64.1} & \textbf{99.2} & 52.7 & 36.4 & \textbf{56.4} & \textbf{75.5} \\
nl & 67.2 & 99.1 & \textbf{77.9} & \textbf{100.0} & 58.5 & 58.8 & \textbf{60.1} & \textbf{87.0} \\
pa & 8.4 & 97.8 & \textbf{23.0} & \textbf{100.0} & 32.1 & 75.5 & \textbf{35.1} & \textbf{80.3} \\
tl & 16.6 & 81.3 & \textbf{60.5} & \textbf{100.0} & 44.4 & 56.7 & \textbf{51.8} & \textbf{87.1} \\
tr & 49.7 & 98.3 & \textbf{69.1} & \textbf{100.0} & 50.6 & 70.7 & \textbf{53.3} & \textbf{92.0} \\
vi & 66.5 & 99.8 & \textbf{79.8} & \textbf{100.0} & 56.9 & 75.4 & \textbf{59.8} & \textbf{94.4} \\
\bottomrule
\end{tabular}
}
\caption{\texttt{SP3F-7B} consistently outperforms Qwen2.5-7B-Instruct even on languages that it was not explicitly trained on. We show languages that were not included in the training set that exist in both tasks.}
\label{tab:unseen_lang}
\end{table}

\subsection{Privileged Information Aids LLM Judges}
We now perform an in-depth exploration of the multiple benefits of privileged information for LLM judge performance. We use $\mathcal{P}_{\mathsf{priv}}$ and $\mathcal{P}_{\mathsf{no-priv}}$ to refer to judges with and without access to reference responses, respectively.

\begin{table}
    \centering
\begin{tabular}{ccc}
\toprule
Input Response-Answer Pair & $\mathcal{P}_{\mathsf{priv}}$ & $\mathcal{P}_{\mathsf{no-priv}}$ \\
\midrule
{\color{expert}\cmark CoT} $\circ$ {\color{expert}\cmark Ans} vs. {\color{error}\xmark CoT} $\circ$ {\color{error}\xmark Ans}  & \textbf{85.77} & 76.42 \\
{\color{expert}\cmark CoT} $\circ$ {\color{expert}\cmark Ans} vs. {\color{error}\xmark CoT} $\circ$ {\color{expert}\cmark Ans} & 77.16 & \textbf{81.08} \\
{\color{expert}\cmark CoT} $\circ$ {\color{error}\xmark Ans} vs. {\color{error}\xmark CoT} $\circ$ {\color{error}\xmark Ans} & \textbf{59.90} & 46.53 \\
\bottomrule
\end{tabular}
    \caption{\textbf{Row 1}: Privileged information increases judge accuracy on responses from Qwen2.5-7B+SFT. \textbf{Row 3}: Access to privileged information increases the judge's accuracy of identifying correct CoT even when the final answer is wrong. This can be important early in RL.}
    \label{tab:judge_type}
\end{table}
\begin{figure}[h]
    \centering
\includegraphics[width=0.7\linewidth]{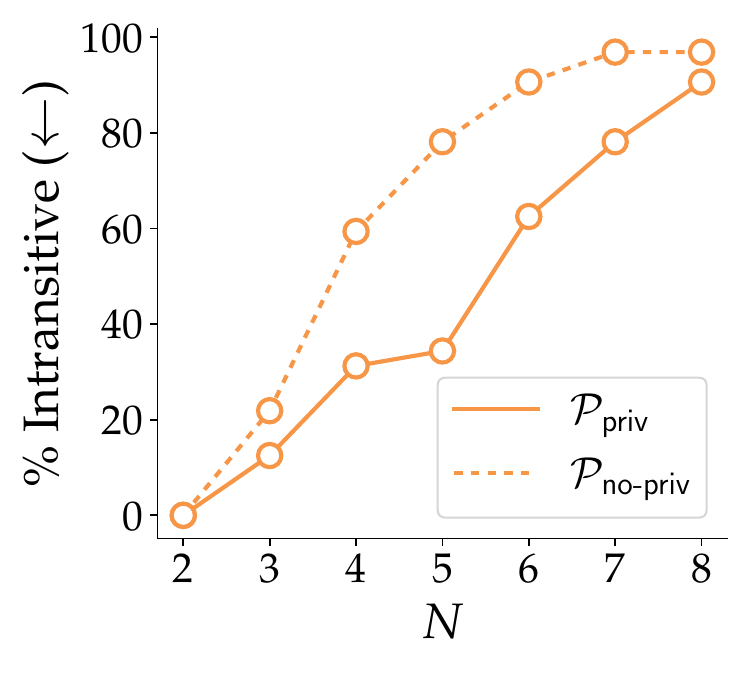}
        \caption{As $N$ increases, it becomes increasingly likely for both privileged and non-privileged judges to have intransitive preferences. However, we consistently find that the privileged judge $\mathcal{P}_{\mathsf{priv}}$ has more transitive preferences than the non-privileged judge $\mathcal{P}_{\mathsf{no-priv}}$. We report the PNT metric proposed by \citet{xu2025investigatingnontransitivityllmasajudge}. We use $N=8$ as the number of training rollouts per sample.}
    \label{fig:intrans}
\end{figure}

\noindent \textbf{Privileged Information Helps With Cold Starts.}
We introduced the pairwise judge reward $r^{\mathcal{P}}$ (Eq. \ref{eq:r-judge}) to provide supervision on the RLM's CoT, especially early on in training when the model may struggle to generate correct final answers. To understand the effect of privileged information on achieving this goal, we first sample responses from the Qwen2.5-7B+SFT model and group them by the correctness of the final answer into correct responses ({\color{expert} \cmark CoT} $\circ$ {\color{expert}\cmark Ans}) and incorrect responses ({\color{error}\xmark CoT} $\circ$ {\color{error}\xmark Ans}). We then graft together different CoTs and answers across correctness groups. Beyond increasing the base accuracy of the judge (\autoref{tab:judge_type}, Row 1), we also observe a significant $13\%$ increase in the ability of the judge model to detect the correctness of the CoT even when the answer is incorrect (\autoref{tab:judge_type}, Row 3). Thus, privileged information appears to help mitigate the early-training cold start issue by increasing the efficacy of CoT supervision. This is particularly important in lower-resource languages where the accuracy of the SFT model is likely to be relatively low.

\noindent \textbf{Privileged Information Reduces Intransitivity.}
While our self-play approach is robust to intransitivity, more consistent preferences can still simplify our learning problem. In \autoref{fig:intrans}, we see a clear reduction in intransitivity when we provide the pairwise judge with privileged information. Thus, beyond merely increasing the accuracy of the judge, privileged information enhances the ability of the judge to provide more consistent \textit{global rankings} (i.e., total orderings) over responses.

\begin{figure}[t]
  \centering
  \includegraphics[width=0.65\linewidth]{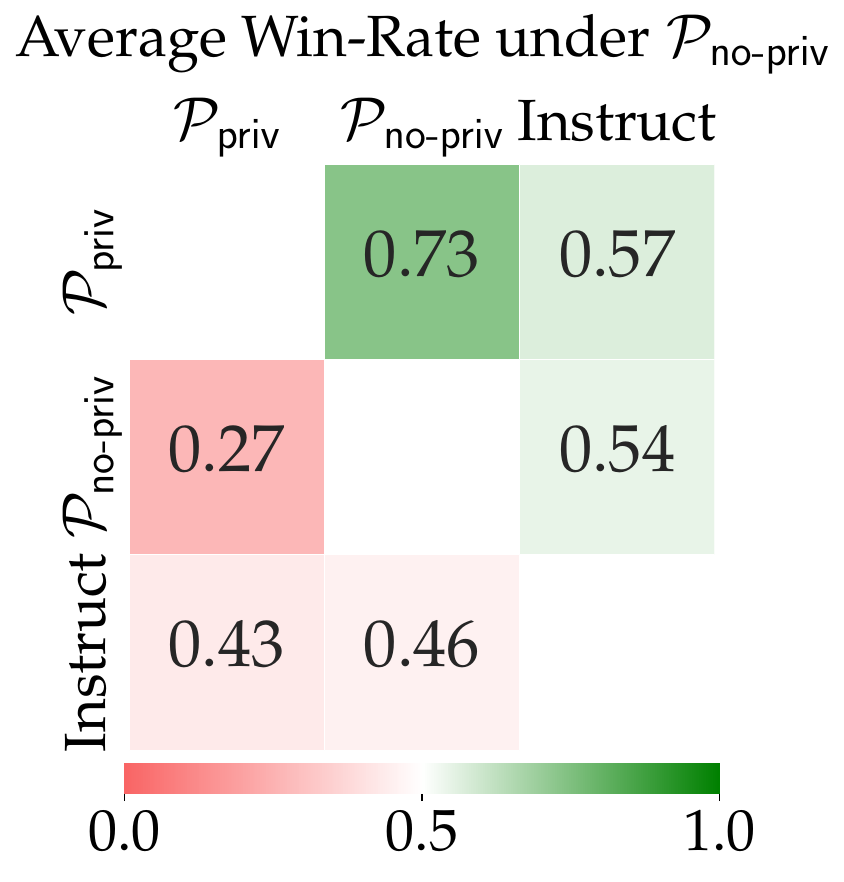}
    \caption{We see $\mathcal{P}_{\mathsf{no-priv}}$ prefer models trained by $\mathcal{P}_{\mathsf{priv}}$ to models trained under its own feedback. Each cell represents the win-rate of the row model against the column model, as evaluated by $\mathcal{P}_{\mathsf{no-priv}}$.}
    \label{fig:winrate}
\end{figure}

\noindent \textbf{Privileged Information Helps Most In-Domain.}
In \autoref{tab:priv_v_nonpriv}, we see models trained with feedback from privileged judges outperform those trained with feedback from non-privileged judges, potentially as a result of the reduction in intransitivity and increase in correct CoT detection capabilities privileged information seems to provide. We see gains on in-domain math tasks and mixed results on out-of-domain non-math tasks.

\noindent \textbf{Models Trained by Judges With Privileged Information Are Preferred Even By Non-Privileged Judges.}
As a final evaluation, we stack the deck in the favor of models trained my $\mathcal{P}_{\mathsf{no-priv}}$ and use $\mathcal{P}_{\mathsf{no-priv}}$ as an oracle to perform model-to-model comparisons. In Figure \ref{fig:winrate}, we see that $\mathcal{P}_{\mathsf{no-priv}}$ prefers models trained via $\mathcal{P}_{\mathsf{priv}}$ to models it trained \textit{itself}, overcoming a \textit{self-preference} bias \citep{panickssery2024llm}. These results potentially indicate that the models trained by $\mathcal{P}_{\mathsf{priv}}$ are qualitatively better models, rather than just narrowly optimizing the preferences of $\mathcal{P}_{\mathsf{priv}}$ in a degenerate fashion.

\begin{table*}[t]
    \centering
    \scalebox{0.85}{\begin{tabular}{l|c|cc|cc}
    \toprule
     &  & \multicolumn{2}{c}{Math} & \multicolumn{2}{c}{Non-Math} \\
    Judge Type & Avg & MGSM & MT Math100 & Belebele & \multicolumn{1}{c}{\makecell{Global MMLU \\ Lite}} \\
    \midrule
    $\mathcal{P}_{\mathsf{priv}}$ & \textbf{64.6} & \textbf{75.3} & \textbf{56.8} & \textbf{72.7} & 53.7 \\
    $\mathcal{P}_{\mathsf{no-priv}}$ & 63.5 & 74.2 & 53.6 & 71.5 & \textbf{54.7} \\
    \bottomrule
    \end{tabular}}
    \caption{
    We see training models with feedback from privileged judges ($\mathcal{P}_{\textsf{priv}}$) performance over those trained with feedback from non-privileged judges ($\mathcal{P}_{\textsf{no-priv}}$). We see particularly strong gains on in-domain math tasks.
    }
    \label{tab:priv_v_nonpriv}
\end{table*}

\section{Related Work}

We provide a brief overview of some related work.

\noindent \textbf{Multilingual Language Modeling.}
Data scarcity is a core problem in multilingual language modeling ~\cite{joshi-etal-2020-state}. A variety of approaches have been proposed to deal with this concern. First, various authors scaled up dataset sizes via Internet scraping ~\citep{xue-etal-2021-mt5, workshop2023bloom176bparameteropenaccessmultilingual}, crowd-sourced high quality data \citep{cahyawijaya-etal-2023-nusacrowd}, manually translated English training data \citep{lai-nissim-2024-mcot, ng2025sealionsoutheastasianlanguages}, or proposed using synthetic training data ~\citep{kautsar2025seadialoguesmultilingualculturallygrounded}. Beyond quantity, quality of data also matters: \citet{zheng2025adamcotrethinkingcrosslingualfactual} propose using rubrics to curate datasets. In parallel, training-free approaches like few-shot prompting ~\citep{Cahyawijaya2024LLMsAF} and representation editing \cite{zhao2025languagemorelanguagereasoningdisentanglement} have elicited multilingual reasoning capability. 

A plethora of techniques have been explored for training multilingual language models. For example, ~\citet{Huang2024MindMergerEBA} merged smaller specifically-trained components into off-the-shelf language models, \citet{zhu-etal-2024-question} trained models to do question translation as a way to improve multilingual performance, and ~\citet{barua2025longchainofthoughtreasoninglanguages} trained models using machine translated or distilled responses from teacher models. Contemporary work has shown how RL training exclusively in high resource languages improves performance on other languages~\citep{huang2025englishcentrictrainingreinforcementlearning}, and explored optimizing language fidelity rewards via RL~\citep{Hwang2025LearnGSA}. In contrast, our work trains on questions in lower-resource languages and goes beyond sparse, verifiable rewards to ease exploration.

\noindent \textbf{Reinforcement Learning Post-Training.} Popularized by RLHF~\citep{ouyang2022traininglanguagemodelsfollow}, reinforcement learning techniques have gained wide adoption in LLM post-training. A wide spectrum of policy optimization algorithms have been proposed, from off-policy regression-based losses \citep{rafailov2023direct, gao2024rebel, azar2024general, gao2024regressing}, to policy gradient techniques \citep{ahmadian2024back, shao2024deepseekmathpushinglimitsmathematical, liu2025understandingr1zeroliketrainingcritical}. We opt for GRPO-style policy gradients in our work due to their relative simplicity and on-policy nature, which fits in cleanly to the self-play algorithmic template \citep{swamy2024minimaximalistapproachreinforcementlearning}.

\noindent \textbf{Supervision via Privileged Information.} Privileged information has provable benefits for reducing the complexity of learning \citep{vapnik2009new}. For example, privileged information has been the key ingredient in recent successes in robotics \citep{choudhury2017datadrivenplanningimitationlearning, chen2019learningcheating, kumar2021rmarapidmotoradaptation, swamy2022sequence, song2025distilldecideunderstandingalgorithmic}. However, because these approaches focus on direct imitation, it is not immediately obvious how to directly apply them in the multilingual reasoning context where we lack sufficiently large amounts of data to imitate in the target language. Thus, we instead provide the privileged information to an LLM judge, similar to the work of \citet{ye2024flaskfinegrainedlanguagemodel, kim2024prometheusinducingfinegrainedevaluation,zhou2025graders}. In particular, we propose providing relatively plentiful English reference answers as privileged information to aid LLM judges in their evaluation of CoTs.

\section{Conclusion}
We introduce \texttt{SP3F}: a data-efficient two-stage framework for improving multilingual reasoning performance without data in target language(s). We find that \texttt{SP3F}-trained models out-perform fully post-trained models across the single language, multilingual, and unseen languages setting while only requiring $\frac{1}{8}$ as much post-training data. We ablate the use of \textit{privileged information} to improve the quality of LLM judgments and find it provides multiple benefits. Thus, an interesting direction for future work is to explore other uses of privileged pairwise judges beyond the multilingual reasoning experiments we perform in our work.

\section*{Contribution Statements}
\begin{itemize}
    \item \textbf{LS} initiated the project, performed all experiments, and wrote the first draft of the paper.
    \item \textbf{GS} came up with the core algorithmic idea of privileged judges, wrote most of the final paper, and helped advise the project.
    \item \textbf{ZSW} and \textbf{GN} advised the project, provided computational resources, and helped with writing.
\end{itemize}

\section*{Acknowledgements} LS and GN were supported in part by a grant from Apple and a compute grant from the CMU FLAME center. GKS and ZSW were supported in part by a STTR grant. We thank Sean Welleck for providing references to other uses of privileged LLM judges and Drew Bagnell for stimulating conversations about the sample complexity benefits of privileged verifiers.

\bibliographystyle{unsrtnat}
\bibliography{ref}

\newpage

\appendix
\onecolumn

\section{Training Hyperparameters}

\begin{table}[H]
\centering
\begin{tabular}{l|c}
\toprule
Hyperparameter & Value \\
\midrule
 \multicolumn{2}{c}{\textbf{Supervised Finetuning}} \\
\midrule
Batch Size & 16 \\
LR & 1e-5 \\
Optimizer & AdamW \\
Training Iterations (Multilingual) & 1000 \\
Training Iterations (Single Language) & 250 \\
\midrule
\multicolumn{2}{c}{\textbf{Reinforcement Learning}} \\
\midrule
Batch Size & 32 \\
LR & 5e-7 \\
Rollouts ($N$) & 8 \\
Sampling Temperature & 1.0 \\
Max Response Length & 2048 \\
$\varepsilon_{\textsf{low}}$ & 0.2 \\
$\varepsilon_{\textsf{high}}$ & 0.28 \\
Training Iterations (Multilingual) & 500 \\
Training Iterations (Single Language) & 250 \\
\bottomrule
\end{tabular}
\caption{We increase the SFT iterations for our multilingual to account for multiple languages. For RL, we use slightly higher $\varepsilon_{\textsf{high}}$ following \citet{liu2025prorlprolongedreinforcementlearning} to encourage more exploration.}
\label{tab:sft_hyperparamater}
\end{table}

\section{Evaluated Language}
\begin{table}[ht]
\centering
\begin{tabular}{ll|ll}
\toprule
\multicolumn{2}{c|}{\textbf{Train Languages}} & \multicolumn{2}{c}{\textbf{Unseen Languages}} \\
\midrule
Code & Language & Code & Language \\
\midrule
ar & Arabic & af & Afrikaans \\
bn & Bengali & nl & Dutch \\
de & German & gu & Gujarati \\
en & English & pa & Punjabi \\
es & Spanish & tr & Turkish \\
fr & French & tl & Tagalog \\
hi & Hindi & he & Hebrew \\
id & Indonesian & vi & Vietnamese \\
it & Italian &  &  \\
ja & Japanese &  &  \\
ko & Korean &  &  \\
pt & Portuguese &  &  \\
ru & Russian &  &  \\
sw & Swahili &  &  \\
te & Telugu &  &  \\
th & Thai &  &  \\
yo & Yoruba &  &  \\
zh & Chinese &  &  \\
\bottomrule
\end{tabular}
\caption{Train and Unseen Languages}
\label{tab:language_list}
\end{table}

\clearpage
\newpage
\onecolumn
\section{Full Table Results}
\label{sec:full_table_results}

In this section, we present per-language evaluation scores for the tasks that we present in the \autoref{tab:all_tasks}. All languages presented here were included in the training dataset. Note that for Translate Test, English is left unevaluated. For the aggregate score, Translate Test is averaged without English.

\begin{table}[ht]
\begin{tabular}{lcccccccccc}
\toprule
 & \multicolumn{10}{c}{MGSM} \\
 & \multicolumn{2}{c}{Avg} & \multicolumn{2}{c}{bn} & \multicolumn{2}{c}{de} & \multicolumn{2}{c}{en} & \multicolumn{2}{c}{es} \\
 & Acc & Lang & Acc & Lang & Acc & Lang & Acc & Lang & Acc & Lang \\
\midrule
Qwen2.5-7B & 22.1 & 90.7 & 2.6 & 97.7 & 27.6 & 88.2 & 40.6 & 100.0 & 28.2 & 83.5 \\
\quad + SFT & 33.7 & 91.4 & 4.7 & 99.2 & 35.4 & 92.7 & 72.5 & 99.9 & 41.6 & 86.9 \\
\quad\quad + RLVR & 65.3 & 99.8 & 53.5 & 100.0 & 75.1 & 99.4 & 91.0 & 100.0 & 83.0 & 99.8 \\
\texttt{SP3F-7B} & 72.5 & 99.4 & 68.2 & 100.0 & 82.7 & 99.7 & 93.1 & 100.0 & 87.5 & 100.0 \\
Qwen2.5-7B-Instruct & 66.4 & 98.4 & 63.0 & 99.4 & 80.3 & 96.8 & 92.2 & 100.0 & 84.2 & 99.3 \\
\quad + Translate Test & 66.2 & 95.8 & 42.9 & 97.2 & 80.2 & 97.5 & N/A & N/A & 88.0 & 100.0 \\
\bottomrule
\end{tabular}

\begin{tabular}{lcccccccccc}
 & \multicolumn{2}{c}{fr} & \multicolumn{2}{c}{id} & \multicolumn{2}{c}{ja} & \multicolumn{2}{c}{ru} & \multicolumn{2}{c}{sw} \\
 & Acc & Lang & Acc & Lang & Acc & Lang & Acc & Lang & Acc & Lang \\
\midrule
Qwen2.5-7B & 45.5 & 79.8 & 31.4 & 91.3 & 21.8 & 92.8 & 26.5 & 94.3 & 0.8 & 85.2 \\
\quad + SFT & 51.5 & 81.2 & 45.6 & 86.8 & 27.2 & 96.2 & 43.0 & 90.8 & 0.6 & 88.2 \\
\quad\quad + RLVR & 78.8 & 99.7 & 79.2 & 99.8 & 64.0 & 100.0 & 81.2 & 100.0 & 12.2 & 100.0 \\
\texttt{SP3F-7B} & 83.6 & 100.0 & 85.0 & 99.5 & 73.1 & 100.0 & 85.8 & 99.9 & 20.5 & 99.7 \\
Qwen2.5-7B-Instruct & 80.2 & 98.8 & 82.3 & 96.8 & 72.5 & 99.8 & 82.5 & 99.5 & 0.8 & 99.0 \\
\quad + Translate Test & 78.0 & 99.8 & 81.8 & 95.1 & 75.2 & 97.2 & 79.7 & 96.9 & 18.0 & 97.8 \\
\bottomrule
\end{tabular}
\begin{tabular}{lcccccc}
 & \multicolumn{2}{c}{te} & \multicolumn{2}{c}{th} & \multicolumn{2}{c}{zh} \\
 & Acc & Lang & Acc & Lang & Acc & Lang \\
\midrule
Qwen2.5-7B & 0.1 & 98.8 & 19.7 & 78.0 & 20.8 & 98.5 \\
\quad + SFT & 0.6 & 99.1 & 33.1 & 75.9 & 48.0 & 99.5 \\
\quad\quad + RLVR & 14.3 & 100.0 & 70.3 & 98.5 & 81.3 & 100.0 \\
\texttt{SP3F-7B} & 34.1 & 100.0 & 72.6 & 94.5 & 83.8 & 99.3 \\
Qwen2.5-7B-Instruct & 13.9 & 99.0 & 62.2 & 92.0 & 82.2 & 100.0 \\
\quad + Translate Test & 29.7 & 99.5 & 73.6 & 73.1 & 80.8 & 100.0 \\
\bottomrule
\end{tabular}
\caption{Evaluation scores per language for Global MGSM}
\label{tab:mgsm_full}
\end{table}

\newpage

\begin{table}[H]
\begin{tabular}{lcccccccccc}
\toprule
 & \multicolumn{10}{c}{MT Math100} \\
 & \multicolumn{2}{c}{Avg} & \multicolumn{2}{c}{ar} & \multicolumn{2}{c}{bn} & \multicolumn{2}{c}{de} & \multicolumn{2}{c}{en} \\
 & Acc & Lang & Acc & Lang & Acc & Lang & Acc & Lang & Acc & Lang \\
\midrule
Qwen2.5-7B & 21.2 & 58.2 & 21.6 & 35.1 & 4.3 & 72.1 & 20.7 & 56.2 & 48.7 & 100.0 \\
\quad + SFT & 26.7 & 58.3 & 25.5 & 36.2 & 8.3 & 67.4 & 27.3 & 61.5 & 56.1 & 99.8 \\
\quad\quad + RLVR & 44.5 & 86.1 & 41.8 & 78.3 & 29.9 & 88.9 & 52.4 & 86.9 & 61.6 & 100.0 \\
\texttt{SP3F-7B} & 56.8 & 82.9 & 57.6 & 72.2 & 48.1 & 74.6 & 62.0 & 92.0 & 67.8 & 100.0 \\
Qwen2.5-7B-Instruct & 52.1 & 65.7 & 45.8 & 51.3 & 47.5 & 56.2 & 61.2 & 65.0 & 66.4 & 100.0 \\
\quad + Translate Test & 60.1 & 59.3 & 58.2 & 46.7 & 57.3 & 48.0 & 65.7 & 63.4 & N/A & N/A \\
\bottomrule
\end{tabular}
\begin{tabular}{lcccccccccc}
 & \multicolumn{2}{c}{es} & \multicolumn{2}{c}{fr} & \multicolumn{2}{c}{hi} & \multicolumn{2}{c}{id} & \multicolumn{2}{c}{it} \\
 & Acc & Lang & Acc & Lang & Acc & Lang & Acc & Lang & Acc & Lang \\
\midrule
Qwen2.5-7B & 25.8 & 57.8 & 33.7 & 64.5 & 9.6 & 51.8 & 23.6 & 52.8 & 28.3 & 48.4 \\
\quad + SFT & 32.1 & 61.2 & 38.0 & 62.2 & 9.8 & 56.6 & 31.6 & 46.8 & 34.5 & 48.9 \\
\quad\quad + RLVR & 51.5 & 98.1 & 53.9 & 92.4 & 36.6 & 84.7 & 49.4 & 79.8 & 51.1 & 84.7 \\
\texttt{SP3F-7B} & 61.9 & 98.9 & 62.5 & 95.5 & 52.8 & 78.3 & 61.1 & 81.1 & 61.9 & 88.8 \\
Qwen2.5-7B-Instruct & 63.3 & 88.6 & 61.2 & 80.3 & 43.8 & 58.8 & 54.7 & 62.0 & 60.0 & 63.9 \\
\quad + Translate Test & 71.6 & 80.0 & 59.6 & 77.7 & 62.9 & 58.2 & 64.9 & 59.5 & 63.6 & 68.4 \\
\bottomrule
\end{tabular}
\begin{tabular}{lcccccccccc}
 & \multicolumn{2}{c}{ja} & \multicolumn{2}{c}{ko} & \multicolumn{2}{c}{pt} & \multicolumn{2}{c}{ru} & \multicolumn{2}{c}{sw} \\
 & Acc & Lang & Acc & Lang & Acc & Lang & Acc & Lang & Acc & Lang \\
\midrule
Qwen2.5-7B & 15.7 & 65.7 & 25.1 & 44.4 & 34.2 & 57.8 & 28.2 & 41.4 & 2.0 & 56.9 \\
\quad + SFT & 21.6 & 57.5 & 26.3 & 44.4 & 41.9 & 61.6 & 33.8 & 35.9 & 2.5 & 63.0 \\
\quad\quad + RLVR & 42.9 & 92.2 & 43.8 & 75.9 & 58.1 & 91.0 & 54.3 & 74.5 & 11.9 & 96.0 \\
\texttt{SP3F-7B} & 57.2 & 95.1 & 59.5 & 76.1 & 62.4 & 96.6 & 63.0 & 77.8 & 27.1 & 77.4 \\
Qwen2.5-7B-Instruct & 53.8 & 80.4 & 55.7 & 43.8 & 63.9 & 83.0 & 59.9 & 46.8 & 7.7 & 67.9 \\
\quad + Translate Test & 67.5 & 73.1 & 56.4 & 40.3 & 67.7 & 81.8 & 63.6 & 45.5 & 32.5 & 68.7 \\
\bottomrule
\end{tabular}
\begin{tabular}{lcccccccccc}
 & \multicolumn{2}{c}{te} & \multicolumn{2}{c}{th} & \multicolumn{2}{c}{zh} \\
 & Acc & Lang & Acc & Lang & Acc & Lang \\
\midrule
Qwen2.5-7B & 1.0 & 86.1 & 17.3 & 29.2 & 20.5 & 63.9 \\
\quad + SFT & 0.8 & 85.5 & 23.6 & 31.1 & 33.6 & 64.5 \\
\quad\quad + RLVR & 12.6 & 97.2 & 37.9 & 46.1 & 55.6 & 91.6 \\
\texttt{SP3F-7B} & 35.9 & 68.2 & 56.3 & 34.7 & 63.1 & 92.7 \\
Qwen2.5-7B-Instruct & 25.1 & 56.3 & 52.9 & 27.5 & 57.6 & 75.0 \\
\quad + Translate Test & 51.5 & 59.9 & 58.2 & 19.1 & 60.23 & 64.39 \\
\bottomrule
\end{tabular}
\caption{Evaluation scores per language for MT Math100}
\label{tab:mt_math100_full}    
\end{table}

\newpage

\begin{table}[H]
\begin{tabular}{lrrrrrrrrrr}
\toprule
 & \multicolumn{10}{c}{Belebele} \\
 & \multicolumn{2}{c}{Avg} & \multicolumn{2}{c}{ar} & \multicolumn{2}{c}{bn} & \multicolumn{2}{c}{de} & \multicolumn{2}{c}{en} \\
 & Acc & Lang & Acc & Lang & Acc & Lang & Acc & Lang & Acc & Lang \\
\midrule
Qwen2.5-7B & 7.5 & 80.4 & 11.9 & 73.7 & 1.6 & 97.0 & 18.4 & 89.7 & 3.7 & 35.6 \\
\quad + SFT & 12.9 & 89.2 & 12.1 & 86.4 & 1.6 & 99.1 & 20.4 & 94.4 & 27.9 & 99.8 \\
\quad\quad + RLVR & 68.2 & 98.7 & 74.6 & 99.8 & 57.8 & 100.0 & 77.8 & 99.9 & 86.8 & 100.0 \\
\texttt{SP3F-7B} & 67.5 & 99.7 & 69.8 & 99.9 & 57.0 & 100.0 & 78.2 & 100.0 & 86.7 & 100.0 \\
Qwen2.5-7B-Instruct & 56.8 & 96.6 & 66.2 & 79.4 & 42.1 & 97.6 & 70.9 & 99.6 & 90.4 & 100.0 \\
\quad + Translate Test & 48.1 & 92.3 & 21.6 & 92.1 & 44.9 & 97.9 & 58.6 & 99.7 & N/A & N/A \\
\bottomrule
\end{tabular}
\begin{tabular}{lrrrrrrrrrr}
 & \multicolumn{2}{c}{es} & \multicolumn{2}{c}{fr} & \multicolumn{2}{c}{hi} & \multicolumn{2}{c}{id} & \multicolumn{2}{c}{it} \\
 & Acc & Lang & Acc & Lang & Acc & Lang & Acc & Lang & Acc & Lang \\
\midrule
Qwen2.5-7B & 3.0 & 91.2 & 15.4 & 93.6 & 0.5 & 90.0 & 11.2 & 91.0 & 11.8 & 94.1 \\
\quad + SFT & 7.9 & 94.7 & 22.8 & 93.9 & 0.6 & 94.7 & 23.9 & 91.8 & 15.2 & 92.3 \\
\quad\quad + RLVR & 80.4 & 100.0 & 81.7 & 99.9 & 55.6 & 99.9 & 76.6 & 99.7 & 78.9 & 99.8 \\
\texttt{SP3F-7B} & 80.4 & 100.0 & 82.7 & 100.0 & 55.7 & 100.0 & 78.3 & 100.0 & 81.5 & 100.0 \\
Qwen2.5-7B-Instruct & 82.5 & 99.8 & 77.8 & 99.9 & 5.9 & 99.8 & 69.0 & 99.5 & 80.8 & 99.8 \\
\quad + Translate Test & 60.2 & 99.4 & 61.1 & 99.5 & 46.3 & 98.5 & 59.4 & 98.8 & 62.5 & 99.2 \\
\bottomrule
\end{tabular}
\begin{tabular}{lrrrrrrrrrr}
 & \multicolumn{2}{c}{ja} & \multicolumn{2}{c}{ko} & \multicolumn{2}{c}{pt} & \multicolumn{2}{c}{ru} & \multicolumn{2}{c}{sw} \\
 & Acc & Lang & Acc & Lang & Acc & Lang & Acc & Lang & Acc & Lang \\
\midrule
Qwen2.5-7B & 2.2 & 73.9 & 23.9 & 79.0 & 11.8 & 87.7 & 6.2 & 72.2 & 0.5 & 82.6 \\
\quad + SFT & 4.2 & 78.3 & 25.3 & 84.8 & 23.4 & 91.0 & 16.7 & 86.6 & 1.4 & 88.8 \\
\quad\quad + RLVR & 74.2 & 99.9 & 76.5 & 99.7 & 79.7 & 100.0 & 80.7 & 100.0 & 35.6 & 99.5 \\
\texttt{SP3F-7B} & 76.6 & 100.0 & 77.1 & 100.0 & 83.0 & 100.0 & 77.4 & 100.0 & 34.8 & 99.9 \\
Qwen2.5-7B-Instruct & 66.1 & 99.2 & 74.9 & 95.2 & 84.5 & 99.9 & 74.3 & 98.8 & 2.4 & 95.0 \\
\quad + Translate Test & 61.3 & 96.8 & 61.2 & 90.2 & 59.8 & 99.8 & 37.7 & 94.7 & 32.9 & 99.3 \\
\bottomrule
\end{tabular}
\begin{tabular}{lrrrrrrrr}
 & \multicolumn{2}{c}{te} & \multicolumn{2}{c}{th} & \multicolumn{2}{c}{yo} & \multicolumn{2}{c}{zh} \\
 & Acc & Lang & Acc & Lang & Acc & Lang & Acc & Lang \\
\midrule
Qwen2.5-7B & 0.3 & 99.3 & 9.5 & 60.9 & 0.5 & 47.5 & 3.0 & 87.9 \\
\quad + SFT & 0.5 & 98.6 & 15.2 & 77.1 & 1.0 & 61.5 & 12.9 & 91.5 \\
\quad\quad + RLVR & 35.4 & 100.0 & 69.6 & 80.0 & 24.5 & 99.1 & 81.0 & 100.0 \\
\texttt{SP3F-7B} & 29.0 & 100.0 & 63.1 & 96.9 & 24.6 & 99.6 & 80.0 & 97.5 \\
Qwen2.5-7B-Instruct & 1.4 & 99.7 & 74.8 & 82.0 & 0.3 & 93.5 & 57.9 & 100.0 \\
\quad + Translate Test & 7.3 & 28.3 & 60.6 & 75.5 & 20.8 & 98.8 & 61.3 & 100.0 \\
\bottomrule
\end{tabular}
\caption{Evaluation scores per language for Belebele}
\label{tab:belebele_full}
\end{table}

\newpage

\begin{table}[H]
\begin{tabular}{lrrrrrrrrrr}
\toprule
 & \multicolumn{10}{c}{Global MMLU Lite} \\
 & \multicolumn{2}{c}{Avg} & \multicolumn{2}{c}{ar} & \multicolumn{2}{c}{bn} & \multicolumn{2}{c}{de} & \multicolumn{2}{c}{en} \\
 & Acc & Lang & Acc & Lang & Acc & Lang & Acc & Lang & Acc & Lang \\
\midrule
Qwen2.5-7B & 8.3 & 85.8 & 8.9 & 72.8 & 2.5 & 97.7 & 17.2 & 91.0 & 18.1 & 100.0 \\
\quad + SFT & 13.5 & 89.6 & 10.6 & 84.1 & 2.2 & 99.4 & 17.8 & 95.5 & 40.3 & 100.0 \\
\quad\quad + RLVR & 53.1 & 99.8 & 52.0 & 99.8 & 38.7 & 99.9 & 62.0 & 99.8 & 70.5 & 100.0 \\
\texttt{SP3F-7B} & 50.8 & 99.5 & 42.9 & 99.4 & 31.8 & 99.3 & 59.1 & 99.8 & 69.5 & 100.0 \\
Qwen2.5-7B-Instruct & 48.2 & 96.2 & 53.5 & 81.9 & 37.2 & 96.9 & 59.4 & 98.2 & 72.9 & 100.0 \\
\quad + Translate Test & 53.7 & 96.5 & 25.2 & 88.9 & 44.8 & 94.8 & 64.2 & 98.4 & N/A & N/A \\
\bottomrule
\end{tabular}
\begin{tabular}{lrrrrrrrrrr}
 & \multicolumn{2}{c}{es} & \multicolumn{2}{c}{fr} & \multicolumn{2}{c}{hi} & \multicolumn{2}{c}{id} & \multicolumn{2}{c}{it} \\
 & Acc & Lang & Acc & Lang & Acc & Lang & Acc & Lang & Acc & Lang \\
\midrule
Qwen2.5-7B & 2.8 & 88.9 & 17.2 & 90.7 & 1.0 & 82.2 & 8.4 & 91.0 & 11.6 & 92.5 \\
\quad + SFT & 7.8 & 92.9 & 21.3 & 90.9 & 1.1 & 92.7 & 17.9 & 87.9 & 13.8 & 92.9 \\
\quad\quad + RLVR & 61.6 & 99.9 & 64.4 & 99.9 & 39.9 & 99.9 & 58.8 & 99.3 & 63.2 & 99.8 \\
\texttt{SP3F-7B} & 59.6 & 100.0 & 61.4 & 100.0 & 34.6 & 99.8 & 56.8 & 99.9 & 61.9 & 99.8 \\
Qwen2.5-7B-Instruct & 66.0 & 99.3 & 62.1 & 99.6 & 7.7 & 98.3 & 57.4 & 98.3 & 66.3 & 98.7 \\
\quad + Translate Test & 68.7 & 98.3 & 66.1 & 99.7 & 45.4 & 97.2 & 67.3 & 96.3 & 69.5 & 98.1 \\
\bottomrule
\end{tabular}
\begin{tabular}{lrrrrrrrrrr}
 & \multicolumn{2}{c}{ja} & \multicolumn{2}{c}{ko} & \multicolumn{2}{c}{pt} & \multicolumn{2}{c}{sw} & \multicolumn{2}{c}{yo} \\
 & Acc & Lang & Acc & Lang & Acc & Lang & Acc & Lang & Acc & Lang \\
\midrule
Qwen2.5-7B & 5.1 & 76.8 & 16.2 & 78.2 & 10.5 & 88.6 & 0.6 & 85.1 & 1.1 & 65.1 \\
\quad + SFT & 8.4 & 83.2 & 18.8 & 83.8 & 23.5 & 92.0 & 1.6 & 83.8 & 1.9 & 74.1 \\
\quad\quad + RLVR & 57.0 & 99.8 & 54.7 & 99.3 & 62.1 & 99.9 & 25.7 & 99.7 & 24.3 & 99.7 \\
\texttt{SP3F-7B} & 58.8 & 100.0 & 57.5 & 99.3 & 61.6 & 100.0 & 26.5 & 99.8 & 19.4 & 99.5 \\
Qwen2.5-7B-Instruct & 58.7 & 98.7 & 59.3 & 91.6 & 67.3 & 99.6 & 3.1 & 97.6 & 1.0 & 85.2 \\
\quad + Translate Test & 65.9 & 96.5 & 53.1 & 87.4 & 64.9 & 99.5 & 31.5 & 97.9 & 19.8 & 98.4 \\
\bottomrule
\end{tabular}
\begin{tabular}{lrr}
 & \multicolumn{2}{r}{zh} \\
 & Acc & Lang \\
\midrule
Qwen2.5-7B & 4.2 & 87.1 \\
\quad + SFT & 15.3 & 91.1 \\
\quad\quad + RLVR & 62.3 & 100.0 \\
\texttt{SP3F-7B} & 60.0 & 95.2 \\
Qwen2.5-7B-Instruct & 50.9 & 99.3 \\
\quad + Translate Test & 65.8 & 99.4 \\
\bottomrule
\end{tabular}
\caption{Evaluation scores per language for Global MMLU}
\label{tab:global_mmlu_lite_full}
\end{table}

\clearpage
\newpage
\section{Prompts}

\subsection{Pairwise Judge Prompts}

\begin{table*}[ht]
\centering
\begin{tabular}{|p{0.9\textwidth}|}
\hline
System Message \\
\hline
\begin{minipage}[t]{0.9\textwidth}
\ttfamily\small
You are an expert judge in evaluating the quality of responses to user queries. \newline
Your task is to determine which response (A or B) is preferable. \newline
You will be provided with the user query and the correct solution. \newline
The responses may be in various languages, but the solution will always be in English. \newline
Decide based on how well does each response align with the correct solution. \newline
The best response should have the closest meaning and intent to the correct solution. \newline
Write your analysis and end it by answering with either \textbackslash boxed\{A\} or \textbackslash boxed\{B\}.
\vspace{0.5em}
\end{minipage} \\
\hline
User Message \\
\hline
\begin{minipage}[t]{0.9\textwidth}
\ttfamily\small
<Query> \newline
... \newline
</Query>\newline
\newline
<Correct Solution>\newline
...\newline
</Correct Solution>\newline
\newline
<Response A>\newline
...\newline
</Response A>\newline
\newline
<Response B>\newline
...\newline
</Response B>\newline
First, using the solution as reference, decide which of the two responses is the closest to the solution. \newline
Finally, choose which is better by answering with either \textbackslash boxed\{A\} or \textbackslash boxed\{B\}. \newline
You MUST provide your reasoning before the answer.
\vspace{0.5em}
\end{minipage} \\
\hline
\end{tabular}
\caption{System and User Prompts for Privileged Pairwise Judge ($\mathcal{P}_{\textsf{priv}}$)}
\label{tab:privileged_info_prompts}
\end{table*}

\begin{table}[H]
\centering
\begin{tabular}{|p{0.9\textwidth}|}
\hline
System Message \\
\hline
\begin{minipage}[t]{0.9\textwidth}
\ttfamily\small
You are an expert judge in evaluating the quality of responses to user queries. \newline
Your task is to determine which response (A or B) is preferable.\newline
You will be provided with the user query and the correct solution. \newline
The responses may be in various languages. \newline
Write your analysis and end it by answering with either \textbackslash boxed\{A\} or \textbackslash boxed\{B\}.
\vspace{0.5em}
\end{minipage} \\
\hline
User Message \\
\hline
\begin{minipage}[t]{0.9\textwidth}
\ttfamily\small
<Query>\newline
... \newline
</Query>\newline
\newline
<Response A>\newline
...\newline
</Response A>\newline
\newline
<Response B>\newline
...\newline
</Response B>\newline
First, decide which of the two responses is preferable. \newline
Finally, choose which is better by answering with either \textbackslash boxed\{A\} or \textbackslash boxed\{B\}. \newline
You MUST provide your reasoning before the answer.
\vspace{0.5em}
\end{minipage} \\
\hline
\end{tabular}
\caption{System and User Prompts for Non-Privileged Pairwise Judge ($\mathcal{P}_{\textsf{no-priv}}$)}
\label{tab:non_privileged_info_prompts}
\end{table}

\clearpage
\newpage
\subsection{System Messages}

\begin{figure}[ht]
    \centering
\includegraphics[width=0.8\linewidth]{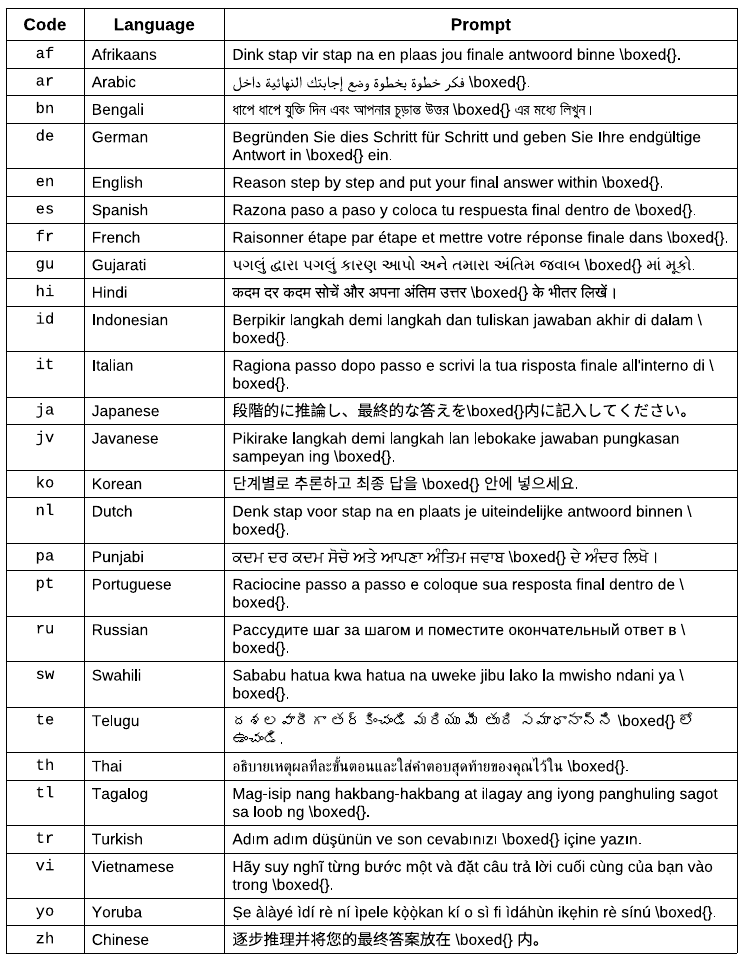}
        \caption{System prompts for all the evaluated languages. Each prompt directs to think step-by-step and write a final answer inside \texttt{\textbackslash boxed\{\}}.}
    \label{fig:reasoning_prompts}
\end{figure}

\end{document}